\documentclass[10pt,twocolumn,letterpaper]{article}

\usepackage{cvpr}
\usepackage{times}
\usepackage{epsfig}
\usepackage{graphicx}
\usepackage{amsmath}
\usepackage{amssymb}

\usepackage[pagebackref=true,breaklinks=true,letterpaper=true,colorlinks,bookmarks=false]{hyperref}

\cvprfinalcopy 

\def\cvprPaperID{****}

\ifcvprfinal\fi

\begin{document}

\title{Scene Recomposition by Learning-based ICP  \vspace{-3 mm}
}

\author{
  Hamid Izadinia\\
  University of Washington\\
  \and
  Steven M. Seitz\\
  University of Washington\\
}

\teaser{
  \includegraphics[width=.96\linewidth]{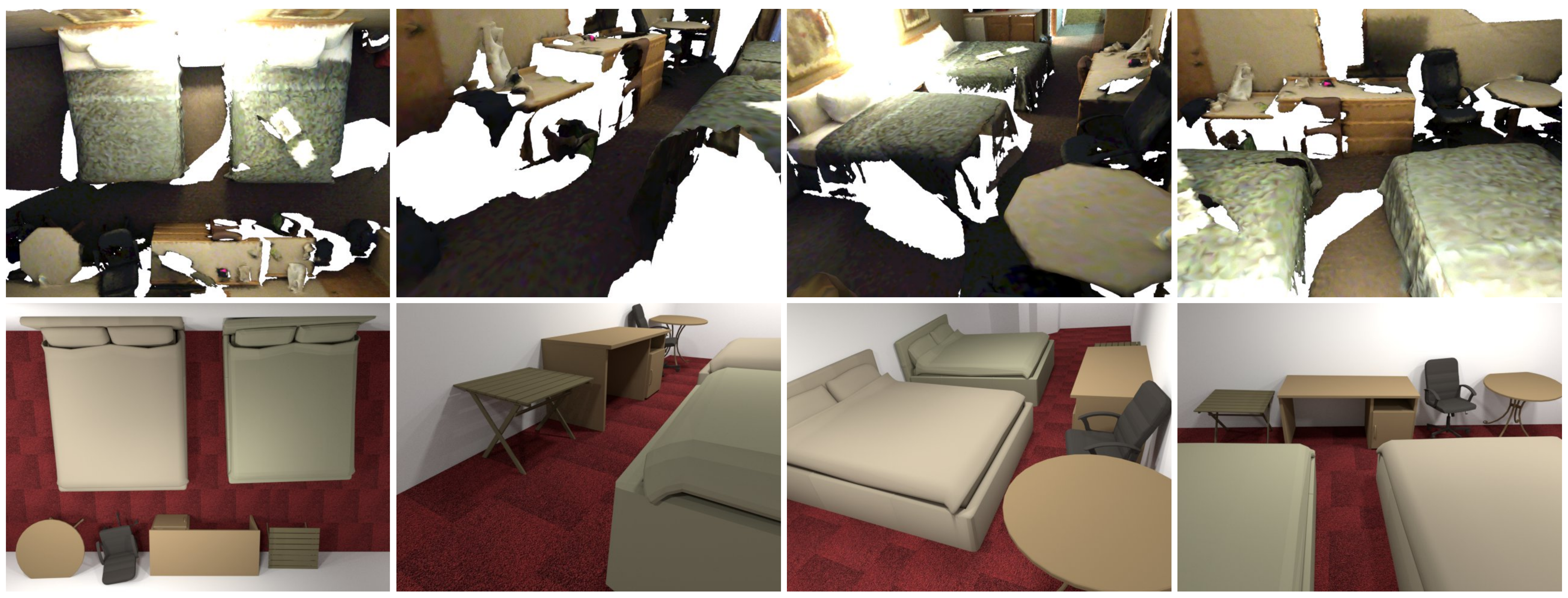}
  \vspace{-.05 in}
   \caption{Given an RGBD sequence from a moving camera, we produce a 3D CAD {\bf \emph{recomposition}} of the scene.  While a fused reconstruction (top) contains holes and noisy geometry, our recomposition (bottom) models the scene as a set of high quality 3D shapes from CAD databases.}
   \label{fig:teaser}
\vspace{-.9 mm}
}

\maketitle

\begin{abstract}
\vspace{-0.1in}
By moving a depth sensor around a room, we compute a 3D CAD model of the environment, capturing the room shape and contents such as chairs, desks, sofas, and tables.  Rather than reconstructing geometry, we match, place, and align each object in the scene to thousands of CAD models of objects.  In addition to the fully automatic system, the key technical contribution is a novel approach for aligning CAD models to 3D scans, based on deep reinforcement learning.  This approach, which we call Learning-based ICP, outperforms prior ICP methods in the literature, by learning the best points to match and conditioning on object viewpoint.  LICP learns to align using only synthetic data and does not require ground truth annotation of object pose or keypoint pair matching in real scene scans. While LICP is trained on synthetic data and without 3D real scene annotations, it outperforms both learned local deep feature matching and geometric based alignment methods in real scenes.  The proposed method is evaluated on real scenes datasets of SceneNN~\cite{scenenn} and ScanNet~\cite{dai2017scannet} as well as synthetic scenes of SUNCG~\cite{song2017semantic}.  High quality results are demonstrated on a range of real world scenes, with robustness to clutter, viewpoint, and occlusion.

\end{abstract}

\vspace{-0.15in}

\vspace{-0.05in}
\section{Introduction}
\label{sec:intro}
\vspace{-0.05in}

3D scene reconstruction is a fundamental challenge of computer vision.  
Most reconstruction techniques focus on estimating surface geometry, in the form of meshes, point-clouds, voxels, or other low-level representations.  Suppose that you had access to a database of 3D models of every object in the world; then you could generate a scene model by identifying which objects are in which locations and placing them there.  We call this variant of the reconstruction problem {\em scene recomposition}.
While previously such an approach was not feasible at scale, the advent of large CAD repositories like {\em ShapeNet}~\cite{shapenet2015} begins to make scene recomposition tractable for real-world scenes.  

Scene recomposition has a number of advantages over scene reconstruction.  First, whereas reconstruction methods often generate holes and capture only visible surfaces, recomposition yields more complete models, including back-facing and hidden geometry~(see Figure.~\ref{fig:teaser}).  Second, CAD models are clean, segmented, and hand-optimized, and thus better suited for applications like games, VR, robotics, etc.  And third, recomposed models can be easily edited by moving objects around, replacing objects, and often come with semantic labels and annotated parts.

Recomposition is not a new idea, dating back to the first ``blocks world'' methods from the 1960s \cite{roberts63}, with a model-based approach to more recent examples of SLAM++ \cite{salas2013slam++} and IM2CAD \cite{izadinia2017im2cad}. We introduce the first fully automatic 3D scene recomposition that takes an RGBD sequence as input and produces a model of the scene composed of best-matched CAD models from {\em thousands} of 3D CAD models.  In addition, we propose a novel learning-based ICP technique for aligning CAD models to scanned geometry.

\begin{figure*}[t]
\centering
\includegraphics[width=.95\linewidth]{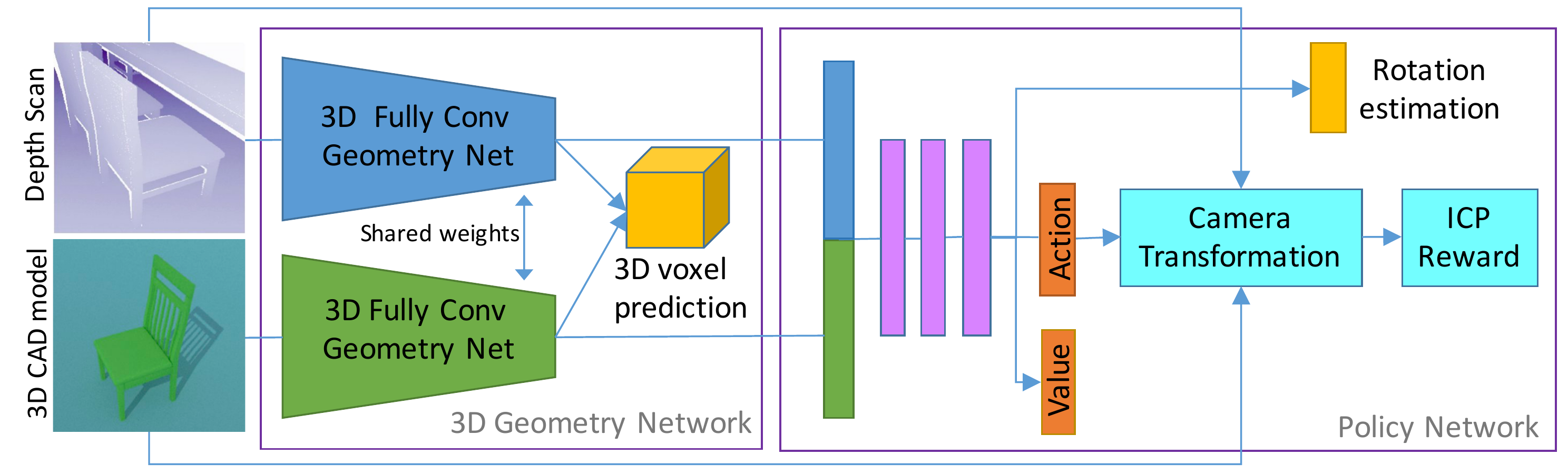}
\vspace{-0.06in}
\caption{\small
   \label{fig:net_arch}
 {\bf LICP Network Architecture}: The input to our network consists of a scanned object paired with a reference CAD model (left) which are processed by the {\bf geometry network} (middle). The geometry network is trained via a supervised loss to predict 3D voxel labels (yellow). The input representations are then concatenated to form the input to the {\bf policy network} (right) which is trained via policy gradient to predict action distribution and value (orange) in order to maximize an ICP reward function. An auxiliary reward function (yellow) that estimates the rotation degree of the 3D CAD model with respect to the scanned shape is also incorporated.}
\vspace{-0.22in}
\end{figure*}

Aligning 3D object models to depth scans is a classical problem in computer vision and geometry processing, and a staple of many practical applications spanning mapping, robotics, and visualization.  The Iterative Closest Point algorithm (ICP) \cite{besl1992method} works by alternating between finding the closest points between the model and the depth image (or other sensor data), solving for the best transformation that aligns the two point sets, and iterating until convergence.  ICP and its variants can robustly converge when the model is initialized close to the solution, but suffer without good initialization or in the presence of significant occlusions and scene clutter.  Matching discriminative local 3D features \cite{spinimage,frome2004recognizing,tombari2010unique,rusu2008aligning,rusu20113d} is an alternative which relaxes the initialization requirements to be more robust, but is less effective for matching synthetic CAD models to real scenes, where 1) the models are simple and feature-poor, and 2) the shapes of the model and real object only approximately agree.

To address these problems, we cast the problem of aligning 3D CAD models to RGBD scans in a reinforcement learning framework which we call Learning-based ICP (LICP).  LICP is trained entirely on synthetic scenes without requiring ground truth annotation of object pose alignment or keypoint pairs in real scenes.  Despite this fact, our quantitative evaluations show that LICP outperforms prior methods in real scenes.  We demonstrate the application of our approach for fully automatic scene recomposition of complex real environments populated with different types of furniture exhibiting a high degree of occlusion.  Our recomposed scenes are comprised of best-matched CAD models from thousands of 3D CAD models in ShapeNet.

\vspace{-0.05in}
\section{Related work}
\label{sec:related}
\vspace{-0.05in}

Inferring 3D object pose and scene recomposition relates to prior works in computer vision and graphics, as follows.

\noindent{\bf ICP:}
ICP was introduced by \cite{chen1992object} and \cite{besl1992method} and solves for the transformation between two point sets. Much research has been devoted to improving this method over the years, including \cite{rusinkiewicz2001efficient,chen1992object,Rusinkiewicz_2019}. Where prior methods focus on feature representation and optimization, we introduce a data-driven and learning-based approach.

\noindent{\bf 3D shape alignment, 3D features and keypoint matching:} An alternative to dense alignment via ICP is to detect robust features (aka {\em keypoints}) to facilitate shape alignment. \cite{spinimage} proposed {\em spin images} and used RANSAC for shape alignment. Other examples of geometric descriptors are Geometry Histograms~\cite{frome2004recognizing}, Signatures and Histograms~\cite{tombari2010unique}, Feature Histograms~\cite{rusu2008aligning} and many more available in Point Cloud Library~\cite{rusu20113d}. However, keypoint methods can be sensitive to noise and do not always perform well, particularly for matching CAD models which are often piece-wise planar and feature-poor. Local features are not robust to symmetries (e.g., all chair legs may have the same features). Model-fitting approaches, also known as registration approaches, try to align an input with a training model but without using descriptors~\cite{besl1992method,jiang2013linear,wu1994recovering}. These approaches do not incorporate learning so that they do not benefit from large amount of data to gain robustness in keypoint detection and matching. Techniques like ~\cite{gupta2015aligning,song2014sliding,geiger2015joint,lai2010object,nan2012search,mattausch2014object,kim2012acquiring} estimate complete scene geometry by fitting instance-level 3D mesh models to the observed depth map. Compared to these methods, our model learns global models over CAD shapes to align poses.

A recent approach for CAD to scan alignment~\cite{Avetisyan_2019_CVPR}, requires \emph{manual} annotation and curation of a large dataset of 3D keypoint correspondences between object CAD model and real scans.~\cite{Avetisyan_2019_CVPR} uses the collected annotation data for learning  correspondences between CAD models and scans. However, our proposed method only uses available synthetic data during training without needing annotated keypoint correspondences in both CAD and real scan domains. While not needing annotated data, our proposed method performs well in the real scene scenarios at the test time. Also to find correspondences at the test time, ~\cite{Avetisyan_2019_CVPR} uses the ground truth object set or a limited set of CAD models, whereas our method can find corresponding CAD models from an \emph{unconstrained} set of objects.

\noindent{\bf Object level RGBD scene reconstruction:} Like our approach, SLAM++~\cite{salas2013slam++} performs room scale semantic object reconstruction using KinectFusion~\cite{newcombe2011kinectfusion} followed by 3D shape recognition. Also, SLAM++ only uses a handful of 3D object models (vs. the thousands in ShapeNet), and does not incorporate a learning-based approach.

\noindent{\bf 3D CAD scene model generation:}
Several prior works proposed methods of generating CAD-based room models using a variety of techniques. Example of these approaches are CAD from text descriptions~\cite{chang2014scenegen},  example based methods~\cite{fisher2012scenesynth} or optimizing furniture arrangements in a space~\cite{yu2011makeithome,merrell2011interactive}. 
Scene models can also be generated by matching 3D objects to a given image~\cite{satkin20153dnn,liu2015model}, rendering a low fidelity synthesize model using RGBD images~\cite{guo2015predicting} or recomposing each scene by analyzing layout and furniture and jointly optimizing their  placements~\cite{izadinia2017im2cad}.

\noindent{\bf Voxel prediction and shape completion:}
Single object shape completion and voxel category prediction has been studied by several authors~\cite{rock2015completing,thanh2016field,wu20153d}. In this paper, we utilize voxel category prediction as an auxiliary loss function to learn 3D representation, but the output of our model is a 3D CAD model with correct pose instead of a voxel grid.  As such, we do shape completion, but compared to prior voxel-wise shape completion methods, our method produces CAD meshes with shape semantics.

\noindent{\bf Shape pose estimation:} Single object 3D pose recognition from a photograph or depth image is also related to our work~\cite{aubry2014seeing,kholgade20143d,salas2013slam++,lim2014fpm,Huang2015,tulsiani2015viewpoints,bansal2016marr,wu2016single}. However our approach differs since we learn the best points to match by conditioning on object viewpoint.

\noindent{\bf Deep feature learning and deep reinforcement learning:} 
A number of researchers have used deep neural networks to learn 3D feature representations ~\cite{song2017semantic,zeng20173dmatch}. Recently, deep Reinforcement Learning (RL) approaches have gained considerable attention due to their success in learning efficient policies to play games~\cite{mnih2015human,silver2016mastering} and obtaining promising performance in robotics~\cite{gu2017deep, andrychowicz2017hindsight}. Part of the success of deep RL is its applicability in solving black-box non-differentiable optimization problems~\cite{sutton2018reinforcement}. Our approach for selecting the correct camera transformation action based on score approximation is closely related to a class of RL techniques called policy gradients~\cite{baxter2001infinite,williams1992simple}. In our method, we have a non-differentiable reward function based on ICP scores of two point clouds and we want to learn the policy that results in receiving maximum reward by using stochastic gradient decent and following a policy gradient update rule.

\vspace{-0.05in}
\section{Proposed Method}
\label{sec:methods}

We begin by describing our learning-based ICP (LICP) approach. Then we explain how to use LICP for recomposing a scene from an input point cloud. For scene recomposition, 3D object detection and 3D semantic segmentation are incorporated for extracting the object instances in the scene. Then, LICP is applied to match and align 3D object CAD models to segmented regions of scene geometry.

\begin{figure*}[t]
\centering
\includegraphics[width=.95\linewidth]{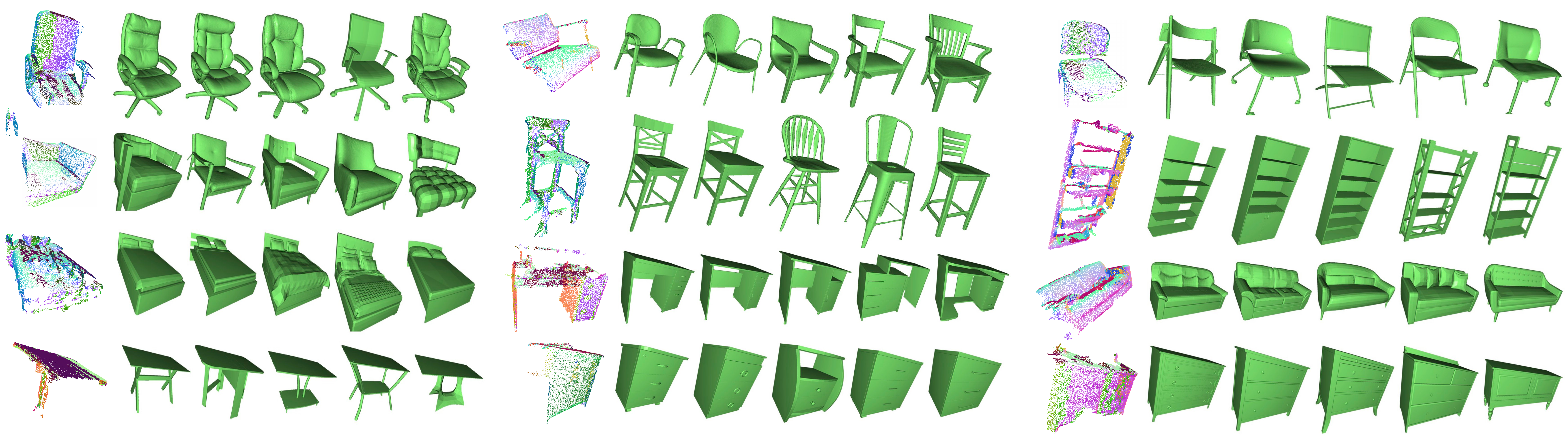}
\vspace{-0.1in}
\caption{\small
   \label{fig:top_retrieved}
Top retrieved CAD models for each object instance segmentation as query. Point cloud query is color-coded with surface normal.}
\vspace{-0.2in}
\end{figure*}

LICP seeks to  estimate the transformation parameters of a scanned rigid object in natural real scenes. This is a challenging task due to inter-object occlusion, self-occlusion and clutter. 
We train a deep neural network that takes in a scanned shape (query) paired with a reference CAD model as input and learns to infer the transformation that should be applied to the reference CAD model to best align its point cloud with the query scan (Figure.~\ref{fig:net_arch}).
To learn such a model, we take advantage of the fact that we can apply any transformation on the reference CAD object and simulate a depth map (point cloud) of the transformed object using ray tracing. To this end, we generate a training set of 3D scans, each paired with a 3D object with known 6DoF parameters. We pose the learning problem in an RL framework where the task is to predict the best action that should be applied to the reference shape such that we can generate the query input scan. Each action encodes a possible 3D transformation that will be applied to the reference 3D shape. By applying each action, we produce a reward that reflects how much the transformed 3D shape matches the query shape.

\subsection{Shape Alignment by Deep RL}
\label{sec:deeprl}
We pose the problem of 3D pose estimation with respect to a reference shape in an RL framework. Suppose we have a reference shape $X^r$ which is presented in a reference pose $P^r$. Using this reference shape, we want to learn to predict the 3D pose of any query 3D object scan $X^q$ that is being cropped out of a complete scene scan. The 3D scan can contain a high amount of occlusion, complicating the alignment process. For representing 3D models, we use a voxel-based 3D feature representation function $\Phi(X)$ for both reference and query shapes. The goal of the RL agent is to select transformation actions to the query object which maximize the expected sum of future rewards. Our reward function, shows the matching score of the query shape with the reference shape if point-to-point local closest point alignment is performed (details in Section~\ref{sec:reward}).

We consider a Markov Decision Process (MDP) defined by states $s\in\mathcal{S}$ and actions $a\in\mathcal{A}$. Each 3D rotational camera transformation is an action $a$ that the RL agent can potentially apply to a 3D shape. 
We define each pair of query object scan and reference object scan captured with camera transformation $\varrho$ as a state $s:(\Phi_{\boldmath{\tau}}(X^q), \Phi_{\varrho}(X^r))$. Each camera transformation action $a$ can transit the agent to a new state by capturing the 3D scan of the reference object $X^r$.
We uniformly discretise the action space of each dimension of rotation degrees into a list of $32$ bins where each bin corresponds to a rotation transformation with a fixed angle. Reducing the action space complexity by discretization accelerates learning and makes it more sample efficient.

\subsection{ICP-based Rewards}
\label{sec:reward}
Each training instance is composed of a 3D point cloud of a scanned query object $\Phi_{\boldmath{\tau}}(X^q)$ captured with an unknown camera pose $\tau$  paired with a reference 3D object $X^r$.
After choosing an action $\boldmath{a}$, we apply the corresponding camera transformation $\boldmath{a}$ and render the transformed point cloud $\Phi_{\boldmath{a}}(X^r)$ of the reference shape $X^r$. Our reward function takes in the point cloud of the query object $\Phi_{\boldmath{\tau}}(X^q)$ and the point cloud of the reference object $\Phi_{\boldmath{a}}(X^r)$ captured under camera transformation imposed by $a$ and produces a score value which reflects how well the two of the point clouds can be matched. We leverage the ICP matching score as the feedback to compute the reward function $f$.  
\vspace{-0.1in}
\begin{equation}
 r(s,a) = f(\Phi_{\boldmath{\tau}}(X^q), \Phi_{\boldmath{a}}(X^r))
\end{equation}
\vspace{-0.2in}

\begin{figure*}[t]
\centering
\includegraphics[width=.97\linewidth]{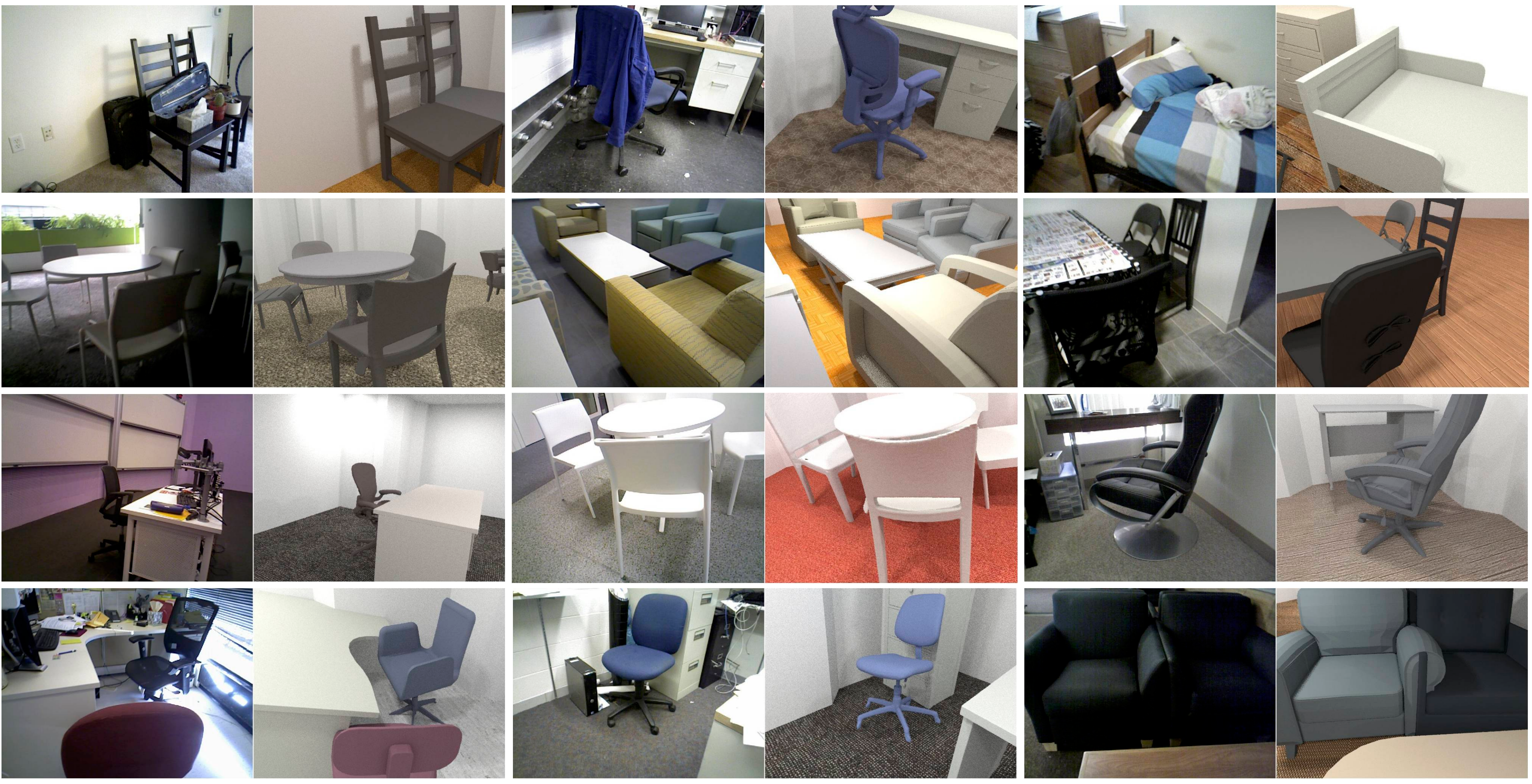}
\vspace{-0.05in}
\caption{\small
   \label{fig:qualitative}
Qualitative examples of the recomposed CAD model of the scene. Each example shows a view of the camera in the scanned scene on \emph{left} and recomposed CAD from the same view on \emph{right}. Our method can successfully recompose cluttered scenes with lots of distractor objects (first row) and huge amount of occlusions in scenes populated with many furniture objects and in confined spaces (second and third Row). Less accurate CAD recomposition can occur due to ambiguous extent of scanned meshes with nearby objects (bottom row, right), or lack of discriminative shape features in different views (cabinet in bottom row, middle) }
\vspace{-0.2in}
\end{figure*}

\subsection{Learning by REINFORCE}
\label{sec:reinforce}
Our reward function is non-differentiable. To solve this black-box optimization problem we opt to use the REINFORCE learning rule~\cite{williams1992simple} where our goal is to find a policy $\pi_{\theta}(a|s)$ with parameters $\theta$ which maximizes the expected sum of rewards: $J(\theta) = \mathbb{E}_{\rho_{\theta}\tau}[R_t]$, where $R_t = \sum_{t}\gamma^{t-1}r(s_t,a_t)$. This expectation is with respect to the distribution of rollout trajectories generated by the policy $\pi_{\theta}$. The gradient of this objective with respect to the parameters $\theta$ can be computed by $\nabla_{\theta}J=\mathbb{E}_{\theta}[\sum_t\nabla_{\theta}\log \pi(s_t|a_t)(R_t - b_t)]$ where $b_t$ is a baseline that does not depend on $a_t$ of the future states and actions. Following a well-known approach, we choose the baseline to be $\mathbb{E}[R_t|s_t]$ and in practice we approximated it with the average value of rewards, updated over time. 

To accelerate training, we augmented the loss function obtained from the REINFORCE learning rule with an auxiliary reward function that is particularly tailored for our task of shape pose estimation. This loss function encodes the error in estimating the rotation angles between the reference CAD model and the shape query scan and corresponds to sum of squared distance between the ground truth rotation and the regressed rotation. 
We use stochastic action sampling based on the probability produced by the current policy. We use dropout~\cite{dropout14, gal2016dropout} to incorporate stochastic action selection and standard epsilon-greedy strategy in RL~\cite{sutton2018reinforcement} for providing exploration in learning.

\subsection{LICP Network Architecture}
Learning a complex shape representation from sparse rewards is very challenging and requires a large number of trials. Instead, we learn the shape representation using dense voxel category labels in a supervised approach, as follows. Freezing the learned shape representation network, we compute features of the 3D observation signal and use a separate network to learn the policy for finding the object poses.

\noindent{\bf 3D Geometry Network:} For 3D geometry feature representation, we use a 3D fully convolutional network that takes in 3D volumes as input and learns to produce per-voxel category labels in a supervised fashion, using softmax loss function over object categories. Each tower of our geometry network uses the 3D fully convolutional architecture of~\cite{song2017semantic} which incorporates several 3D convolution layers.

\noindent{\bf Input volume generation:} Our observation signal is in the form of 2D depth maps, which we convert to a volumetric grid of Truncated Distance Function (TDF) values. The TDF representation can encode both single depth and multiple depth images.
Specifically, each voxel takes a value which indicates the distance between the center of that voxel to the nearest 3D surface. Following~\cite{zeng20173dmatch}, these values are truncated, normalized and then inverted to be between $1$ and $0$, indicating on surface and far from surface, respectively.

\noindent{\bf Policy Network:} Our policy is learned via a fully connected network consisting of three layers, each with 256 units followed by dropout and ReLU, using the policy learning and loss and reward function in Sections~\ref{sec:reward} and~\ref{sec:reinforce}.

\noindent{\bf Training Details:} We implement our model in TensorFlow~\cite{abadi2016tensorflow} and use stochastic gradient descent with a learning rate of $0.001$ and decay factor of $0.95$. We train both 3D geometry and policy network over more than 1 million training samples in simulation.

\subsection{Generate Training Data using Simulation}
\label{sec:data_gen}
We generate synthetic training data using SUNCG scenes~\cite{song2017semantic}. 
In each room, we move the camera at a person's height while looking at different objects in the scene. 
We generate a wide range of camera angles: yaw varies between $[-180,180]$, pitch depends on the height of objects and varies between $[-90,90]$ and roll randomly takes a value in $[-10,10]$ degrees. To produce a variety of viewpoints, we jitter the camera with a small amount of noise. For each view, we capture the depth image and crop the box around the object which also contains some parts of the other objects. We then pass the partial point cloud to the network as input. We rasterize the mesh of the 3D CAD model into a point cloud and use the produced point cloud as the reference input of the network. The truncated distance function of the point cloud is used as input to the network.

\subsection{Scene Recomposition}
Our scene recomposition pipeline takes in a point cloud which is produced from RGBD video of a real scene. 
We apply 3D object detection and semantic segmentation for extracting 3D object instances. Then, we use the output of our trained 3D geometry network (see Figure~\ref{fig:net_arch}) for finding the nearest 3D CAD model in the set of CAD models and use it as reference 3D shape. Finally, we deploy LICP for aligning the 3D CAD model to object instance segmentation, as described in Section~\ref{sec:deeprl}.

\begin{figure*}[t]
\centering
\includegraphics[width=.99\linewidth]{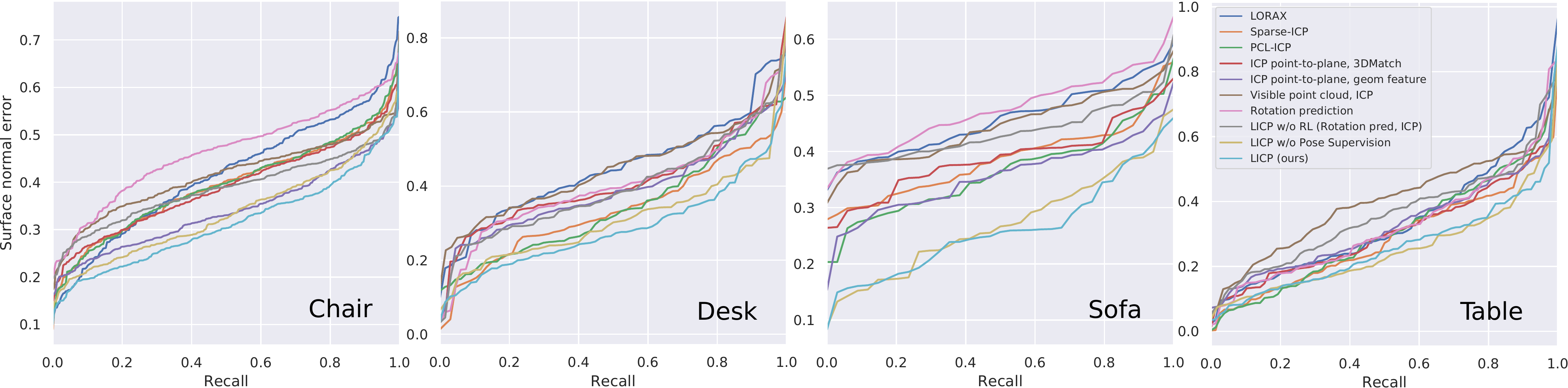}
\vspace{-0.05in}
\caption{\small
   \label{fig:quantitative_localfeat}
   Comparison of proposed LICP method with local feature matching and alignment methods on real data (\emph{lower values are better}). The legend is only shown on the right plot for better readability and the color of methods are the same for all plots.}
\vspace{-0.2in}
\end{figure*}

\noindent{\bf 3D Object Detection: }
We use the two-step object detection regime~\cite{ren2015faster,chen2017multi, ku2018joint} as follows. We train a category agnostic region proposal network which gives the objectness score for different 3D bounding boxes over the point cloud. We simultaneously train another network for classification of 3D bounding boxes for each of the object categories. Both networks share the feature extraction layers which are based on the VGG architecture~\cite{vgg}. We use cross entropy loss for both region proposal and classification networks. We also learn the deviation of the 3D boxes using regression loss in $x$ and $y$ dimensions and the $z_l$ and $z_h$ for the lower and higher extent of the object along the $Z$ axis orthogonal to the ground plane. We rectified the point cloud in  world coordinates by rotating  the gravity direction and then making it axis aligned with the dominant $X$-$Y$ orientation on the ground plane. To compute feature maps from the point clouds we use the orthographic projection of the point cloud representations and extract feature from planes in different heights following~\cite{chen2017multi}. For training, we use rendered depth images from SUNCG~\cite{song2017semantic} as explained in~\ref{sec:data_gen}. We use the entire scene composed of multiple objects in the field of view for each camera pose. We set $0.5$ as the threshold for intersection over union (IoU) of 3D detection boxes and use non-maximum suppression for removing low scoring 3D boxes which have high overlap with higher scoring detections. We find the translation and scale of the objects via 3D object detection and apply the inferred translation and scale to the CAD models.

\noindent{\bf 3D Semantic Segmentation: }
Clean object instance segmentation is important for the alignment stage of our method. For instance, when a chair is next to a table the 3D bounding box of the chair may include some part of table and vice versa. In order to remove such distractors from the detection bounding box of each object detection we incorporate the semantic segmentation inferred on the point clouds. We take all points inside the 3D detection box and remove the points with semantic label of other object categories with overlapping detected bounding boxes. We also remove the points with ``floor'' and ``wall'' labels. We follow~\cite{pointnetseg,LiPoint,pointnetplusplus} for training semantic segmentation over the point cloud and learn a model for all object categories as well as floor and wall classes.

\noindent{\bf Room Layout Estimation and Scene Visualization: }
We use the inferred wall points from the 3D point cloud segmentation to estimate the room layout. For each point on the ground plane ($X,Y$), we count the number of wall 3D points, aggregating over the $Z$ axis. The locations on the ground plane with high frequency of wall voxels define the boundary of the room. We use the extent of the floor voxels wherever scan does not have wall in the boundary. 
Once all wall voxels on the ground plane are computed we run the concave hull algorithm to find the room boundary. We infer the location of the floor plane to be at the $Z$ which has the highest frequency of floor voxels inferred via semantic segmentation of 3D points. The color of each object is estimated by medoid color of the point clouds belonging to the object instance segmentation. The floor texture is selected based on the feature similarity to a set of texture image.


\vspace{-0.05in}
\section{Experiments}
\label{sec:experiments}
\vspace{-0.05in}
In our experiments we want to investigate: 1) How accurate is our learning-based ICP compared to non-learning previous approaches,  2) how does our method compare with keypoint matching approaches based on deep features, and 3) how can our model be applied in scene CAD model recomposition of unstructured and cluttered real world environments.  To answer these questions, we evaluate the performance of our method both quantitatively and qualitatively. For real-world evaluation, we use the publicly available SceneNN~\cite{scenenn} and ScanNet~\cite{dai2017scannet} datasets. SceneNN and ScanNet test sets contain scans of 95 and 312 scenes from different real world indoor spaces, respectively. These scene point clouds are scanned from various offices, bedrooms, living room, kitchen, etc., and exhibit a diverse collection of unstructured real world scenes populated with various furniture types, styles, and types of clutter from many distractor objects. These scenes are scanned with commodity depth cameras and we use the fused output.

\begin{figure*}[t]
\centering
\includegraphics[width=.99\linewidth]{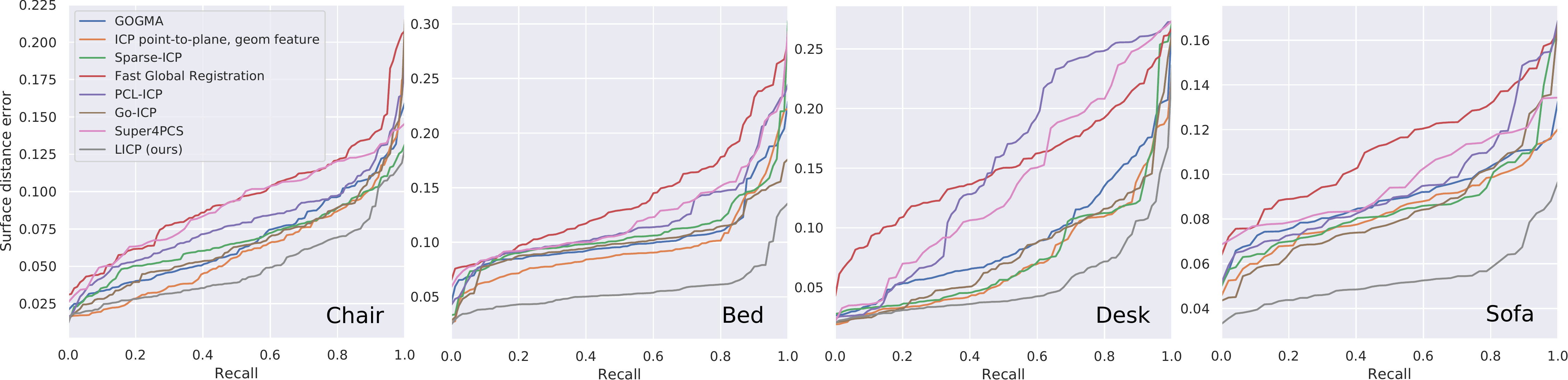}
\vspace{-0.05in}
\caption{\small
   \label{fig:sim_quant_alignments}
   Comparison of proposed LICP method with robust and global alignment algorithms on synthetic data (lower values are better).}
\vspace{-0.18in}
\end{figure*}

\subsection{Quantitative Evaluation}
\label{sec:quant}
We evaluate the accuracy of our method for 6DoF pose estimation of furniture objects in both real and synthetic scenarios. We compare our results with prior works of~\cite{chen1992object, rusu2009fast, zeng20173dmatch, lorax, sparseicp, zhou2016fast, yang2016go, campbell2016gogma, mellado2014super}. For the evaluation criteria, we compute the alignment error between the scanned mesh and the CAD model with the predicted pose. To compute the alignment score, the closest point on the CAD model is found for each point in the input scan and the cosine distance between surface normals is computed. In the synthetic data experiment, we use the distance between points on reference CAD model and scan given that we have access to the ground truth mesh of the object in simulation.

\noindent{\bf Quantitative evaluation on real data:} To evaluate the effectiveness of LICP for 6DoF object pose estimation, we incorporate the ground truth point cloud segments and object labels. We use the feature representation of our trained 3D geometry network for finding the nearest 3D CAD model from a database of 1550 CAD models from~\cite{song2017semantic,pix3d} and use it as the reference CAD model. The quality of the object style match for retrieved CAD models is shown for several examples in Figure~\ref{fig:top_retrieved}.

We compare LICP with local feature matching and variants of ICP from the literature. For local feature matching, we compare against the hand-designed geometric feature of FPFH~\cite{rusu2009fast}, learned local deep feature by 3DMatch~\cite{zeng20173dmatch} and LORAX~\cite{lorax}. After matching the local features, we use RANSAC for coarse registration followed by point-to-plane ICP~\cite{chen1992object} for fine alignment of CAD model and input scan. For comparing against LORAX, we use the released code of~\cite{lorax} for super-point extraction and use local deep features learned in an unsupervised fashion from point clouds of synthetic object CAD models via GAN. We also compare with \emph{Sparse ICP}~\cite{sparseicp} (a variant of ICP that is robust to input noise), and the \emph{PCL} implementation of ICP. Figure~\ref{fig:quantitative_localfeat} summarizes our quantitative comparison results. In the plots of Figure~\ref{fig:quantitative_localfeat} \emph{``ICP point-to-plane, geom feature''} refers to FPFH setting. As demonstrated in Figure~\ref{fig:quantitative_localfeat}, our method outperforms all aforementioned prior methods.

We also compare LICP with other baselines and variants of proposed LICP with different combinations of loss and reward function. {\bf Rotation prediction} only uses object rotation estimation output of the learned network in Figure~\ref{fig:net_arch} and does not use our RL component. {\bf Rotation pred., ICP point-to-plane} uses the rotation estimation output of the LICP network and applies ICP point-to-plane for finer object alignment. {\bf Visible point cloud, ICP} only uses the visible points of the point cloud from predicted object pose for ICP alignment. {\bf LICP w/o Pose Supervision} uses a policy network that is only trained with RL component and without strong object pose supervision of auxiliary loss. All of these variants have lower performance than our full LICP model that combines ICP-based reward and auxiliary loss for learning the policy network. Also the performance of LICP only with RL is close to LICP which suggests that LICP performance is mostly gained by RL learning rather than strong object pose supervision.

We do not have access to the ground truth CAD model of the shapes in the input scan and we use the surface normal error between recomposed CAD and input scan. We plot the surface normal error vs. recall for each category, which is the percentage of samples with surface normal error lower than each error value. Note that the smallest average ICP distance between the pair of scan and CAD model never goes to zero since the point cloud input pairs to the ICP method are sampled differently and are never identical.

\noindent{\bf Quantitative evaluation on synthetic data:} 
We test on the SUNCG~\cite{song2017semantic} test set where objects are placed in 3D scenes with realistic furniture arrangements. This experiment is performed on several input CAD models and input scans. The alignment error is the mean surface point distance in meters between the object surface in scan and the reference CAD model. In this experiment we test on synthetic scans where we have the ground truth surface of the scanned object. Therefore, we can compute the distance between the surface of the reference CAD and surface of the CAD in the scan. We compare LICP with robust and global alignment algorithms: Fast Global Registration~\cite{zhou2016fast}, globally-optimal algorithm Go-ICP~\cite{yang2016go}, GOGMA~\cite{campbell2016gogma}, Super4PCS~\cite{mellado2014super} and Sparse ICP~\cite{sparseicp}. We also compare LICP against point-to-plane ICP~\cite{chen1992object} with FPFH geometric feature and PCL implementation of ICP. The results are summarized in Figure~\ref{fig:sim_quant_alignments}. Our LICP alignment outperforms other global and robust alignment methods by a large margin.

\begin{figure}[t]
\centering
\includegraphics[width=\linewidth]{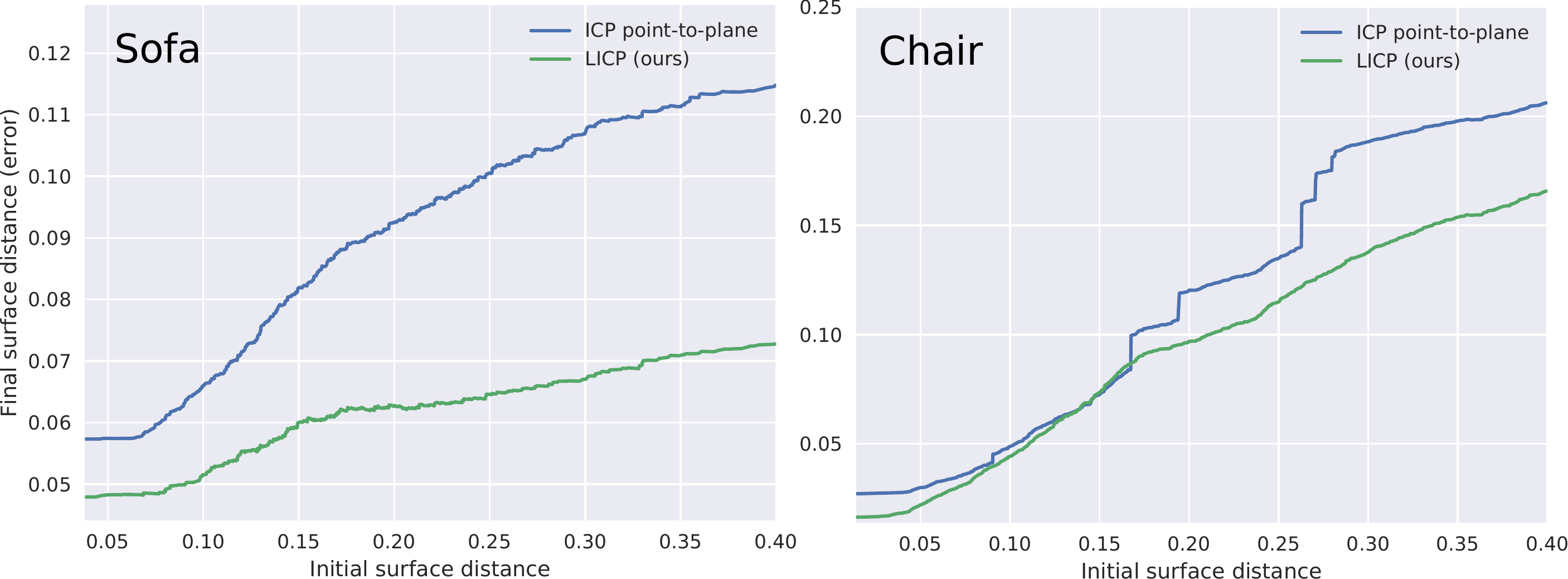}
\vspace{-0.15in}
\caption{\small
   \label{fig:synth_quant}
Evaluating the robustness of our proposed LICP method for aligning 3D CAD models with drastic orientation differences to the input scan using synthetic data.}
\vspace{-0.25in}
\end{figure}

We also evaluate the robustness of LICP against large orientation differences between the object scan input and the reference CAD model and compare against Chen and Medioni ICP~\cite{chen1992object} in Figure~\ref{fig:synth_quant}. The reference CAD models are initialized with different orientations for each experiment. In Figure~\ref{fig:synth_quant}, the x-axis shows the initialization error while the y-axis shows the final alignment error after ICP is converged. While both methods reduce the alignment error, LICP obtains lower final error compared to \cite{chen1992object}.

\begin{figure*}[t]
\centering
\includegraphics[width=.97\linewidth]{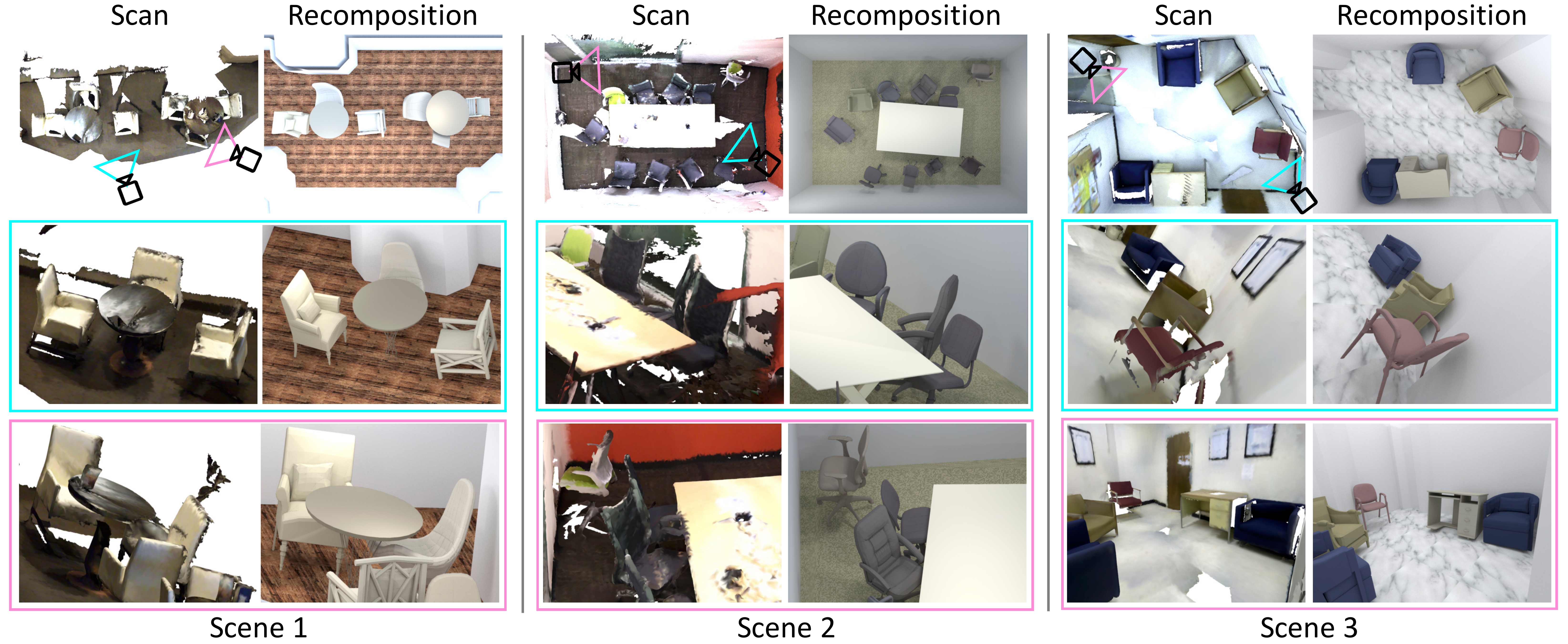}
\vspace{-0.05in}
\caption{\small
   \label{fig:scene_recomp_fig}
  Scene \emph{recomposition} using our proposed fully automatic method. Scene recomposition is shown for three different scenes. In each scene, the top row shows the top-down view of the scene; the middle and bottom rows demonstrate two close-up views of each scene. Camera location and pose is color coded on top-down view). }
\vspace{-0.22in}
\end{figure*}

\subsection{Qualitative Evaluation}
\label{sec:sim}
\noindent{\bf Real scene shape alignment: } 
Figure~\ref{fig:qualitative} demonstrates several examples of scene CAD models recomposed (on right) from the depth scan of real scenes (on left) by applying our algorithm where best-matched CAD models and 6Dof object poses are estimated. The first row in Figure~\ref{fig:qualitative} shows several recomposed CAD scene models in the presence of a high amount of scene clutter. For example, the surface of the two chairs on the top left is filled with random objects, and the back cushion of the blue office-chair (first row, middle figure) is occluded with a shirt. While such arbitrary objects results in significant amount of noise in the depth scans, our method can estimate the 6DoF pose and object style reasonably well. Examples of the second row in Figure~\ref{fig:qualitative} show scenarios with significant occlusions as the result of a densely populated scene.
As shown in the figure, our method handles such occlusions well and  produces CAD scene models with accurate object pose and styles. Several failure cases are shown in the bottom row of Figure~\ref{fig:qualitative} where the estimated object poses are less accurate. For example, in the middle example of the forth row, the pose of the cabinet behind the blue chair is not estimated correctly due to the lack of strong discriminative shape features between the right face and the front face of the cabinet. Also the retrieved armchair style is not accurate in the left example of the forth row, as the extent of the armchair cannot accurately be obtained from the scanned point cloud because of high level of occlusion with the nearby objects. 

\vspace{-.01in}
\noindent{\bf Real scene recomposition: }We deploy our fully automatic scene recomposition method on real scenes, with results shown in Figure~\ref{fig:scene_recomp_fig}. For each scene, we render two different close-up camera viewpoints and the top-down view of the scene recomposed by our method and also show corresponding views from the scan. As shown in Figure~\ref{fig:scene_recomp_fig}, these scenes are densely populated with different furniture and the scene scans contain many holes. Despite many occlusions and holes, our method produces satisfying scene recompositions.
Using TITAN Xp GPU, the computational time for a typical scene with an average complexity is approximately 6.5 seconds for 3D object detection and 9.5 seconds for 3D semantic segmentation. LICP 3D CAD alignment takes 1.22 seconds per object instance which includes 0.65 seconds for 3D Geometry Net, 0.008 seconds for Policy Net and 0.56 seconds for ICP Reward. 

\noindent{\bf Surface point visualization during inference: }
\label{sec:vis_model}
LICP learns to assign different weights to surface points of the reference CAD model when queried with arbitrary posed object scans. The assigned weights for surface points in the reference CAD model are computed based on the visible surface points. The visible surface points are captured via ray tracing from the actions inferred, i.e., the camera transformation multiplied with the value estimated by the value function in our policy network. These weights reflect the contribution of each surface point in inferring the correct transformation action.

\begin{figure}[t]
\centering
\includegraphics[width=\linewidth]{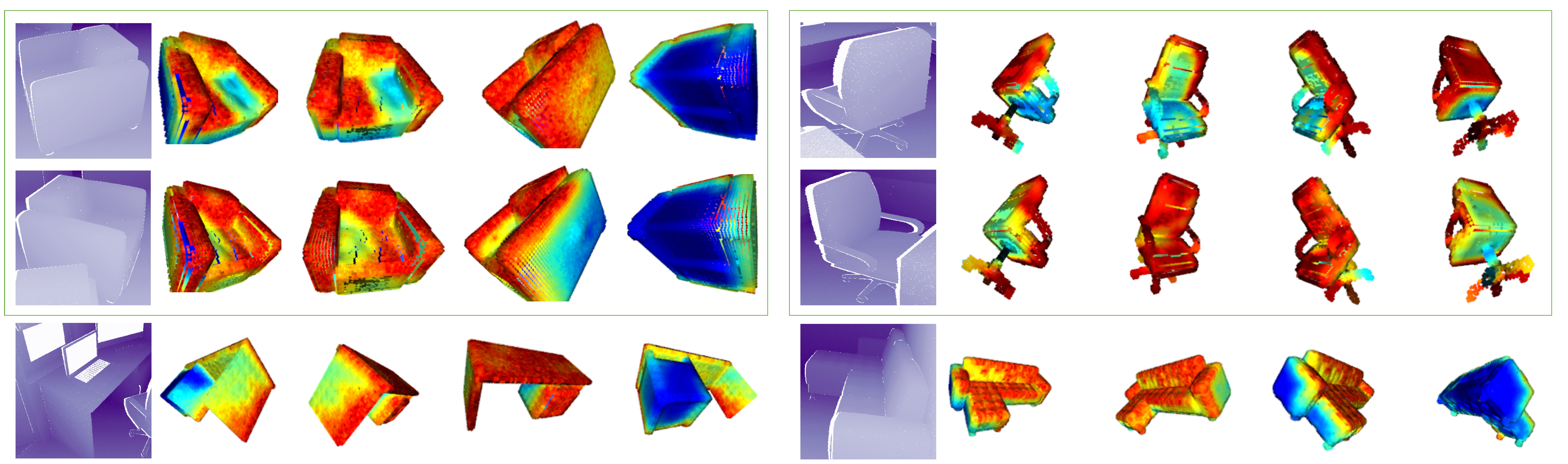}
\caption{\small
   \label{fig:weight_vis}
Visualization of the learned weights (\emph{right}) for different samples and various query scan viewpoints (\emph{left}). The learned weights are shown from four different views of the reference CAD model. Weight values are color-coded from low (\emph{blue}) to high (\emph{red}). The first two rows show that the surface points of the same reference CAD model are assigned with different weights depending on the query scan viewpoint.}
\vspace{-0.18in}
\end{figure}

Figure~\ref{fig:weight_vis} shows the surface point weights obtained for different objects when queried with scans from various viewpoints. The assigned weights are conditioned on the viewpoint of the query shape. When LICP is queried with a left-sided armchair, the visible surface points on the left side of the reference armchair gain higher weights and vice versa. Similarly, office chairs with different poses and occlusion patterns are provided. LICP assigns higher weights to the surface points that are not occluded and ignores the contribution of the occluded surface points. 
The bottom row of Figure~\ref{fig:weight_vis} shows similar patterns in the produced weights for surface points of desk and L-shaped sofa instances.

\vspace{-0.07in}
\section{Conclusion}
\label{sec:conclusion}
\vspace{-0.1in}

In this paper, we compute 3D scene recompositions from a sequence of RGBD scans captured by a moving camera from a real scene. We present a learning based approach for shape alignment called Learning-based ICP (LICP). LICP combines deep 3D feature learning with reinforcement learning and is able to infer the 6DoF object transformation with respect to a reference shape. By leveraging large scale shape 3D databases and learning the transformation policy for various object poses, LICP becomes robust to scene clutter and partial occlusions.  Our experimental results on diverse real world scans demonstrate high performance of our method compared to various baselines.

\vspace{-0.13in}
\subsubsection*{Acknowledgments}
\vspace{-0.05in}
This work was supported in part by the University of Washington Animation Research Labs and Google.

{\small
\bibliographystyle{ieee}
\bibliography{LICP}
}

\end{document}